\documentclass[leqno]{article}
\usepackage{amsmath,amsthm,amssymb,named,proceedings}

\newtheorem{thm}{Theorem}[section]
\newtheorem{cor}[thm]{Corollary}
\newtheorem{lem}[thm]{Lemma}

\theoremstyle{definition}
\newtheorem{defn}{Definition}[section]
\theoremstyle{remark}
\newtheorem*{rem}{Remark}

\newtheorem{exam}{Example}[section]
\newcommand{\B}[1]{\mathbf{#1}}
\newcommand{\BB}[1]{\mathbb{#1}}
\newcommand{\C}[1]{\mathcal{#1}}

\DeclareMathOperator{\Th}{Th} 
 
\begin{document}

\title{Two Representations for\\ Iterative Non-prioritized Change}
\author{{\bf Alexander Bochman}
\\ \emph{Computer Science Department}
\\ \emph{Holon Academic Institute of Technology, Israel}
\\ \emph{email}: bochmana@hait.ac.il}

\date{}

\maketitle
\begin{abstract}
We address a general representation problem for belief change, and describe two
interrelated representations for iterative non-prioritized change: a logical
representation in terms of persistent epistemic states, and a constructive representation
in terms of flocks of bases.
\end{abstract}

\section{Introduction}

Two main kinds of representation for belief change have been suggested in the literature.
In the AGM theory \cite{AGM85,Gar88}, belief change is considered as a change of
\emph{belief sets} that are taken to be deductively closed theories. According to the
so-called base approach, however, a belief set should be seen as generated by some
(finite) \emph{base} (see, e.g., \cite{Han99}). Consequently, revisions of belief sets
are determined on this approach by revisions of their underlying bases. This drastically
reduces the set of alternatives and hence makes this approach constructive and
computationally feasible.

The AGM theory and base approach constitute two seemingly incomparable representations
for belief change, each with advantages of its own. Still, the framework of epistemic
states suggested in \cite{EPI,BOOK} allows to formulate both as different species of a
single general representation. The reformulation makes it possible, in particular, to
reveal common constitutive principles behind them. These common principles embody,
however, also common shortcomings that lead to a loss of information in iterated changes.
This suggests epistemic states in their full generality as a justified representational
alternative.

Epistemic states could be primarily seen as a generalization of the AGM models, namely,
they form an abstract logical representation for belief change. Though the base-oriented
models can be translated into epistemic states, the translation loses some important
advantages of such models, namely their inherent finiteness and constructivity. Hence it
would be desirable to have a constructive representation for epistemic states that would
preserve these features of base models while avoiding their problems. Fortunately, such
an alternative representation has already been suggested in \cite{FUKV86}, and it amounts
to using sets, or \emph{flocks}, of bases. As we will show, a certain modification of the
original flock models will provide us with a constructive representation for an important
class of epistemic states and belief change processes in them.

\begin{rem}
The theory described below, is different from that suggested in \cite{BASE,AGM}, since
the latter has turned out to be flawed in its treatment of expansions. The shortcoming
has created the need for shifting the level of representation from pure epistemic states
used in the above papers to more general persistent epistemic states (see below).
\end{rem}

\section{Epistemic states}

A common feature of the AGM and base representations is a preference structure on certain
subsets of the belief set. In the case of the AGM paradigm, it is a preference structure
on the maximal subtheories of the belief set, while for the base representation it is a
preference structure on subtheories that are generated by the subsets of the base. The
notion of an epistemic state, defined below, reflects and generalizes this common
structure of the two representations.

\begin{defn}
An \emph{epistemic state} $\BB{E}$ is a triple $(\C{S},l,\prec)$, where $\C{S}$ is a set
of \emph{admissible belief states}, $\prec$ is a preference relation on $\C{S}$, while
$l$ is a labelling function assigning a deductively closed theory (called \emph{an
admissible belief set}) to every admissible belief state from $\C{S}$.
\end{defn}

Any epistemic state determines a unique set of propositions that are \emph{believed} in
it, namely, the set a propositions that hold in all maximally preferred admissible belief
states. Thus, even if an epistemic state contains conflicting preferred belief states, we
can still believe in propositions that hold in all of them. Still, the belief set
associated with an epistemic state does not always constitute an admissible belief set by
itself. The latter will hold, however, in an ideal situation when an epistemic state
contains a unique most preferred admissible belief state. Such epistemic states will be
called \emph{determinate}. Both AGM and base models correspond to determinate epistemic
states. Nevertheless, non-determinate epistemic states have turned to be essential for an
adequate representation of both nonmonotonic reasoning and belief change. In particular,
the necessity of accepting non-determination arises most clearly in the analysis of
contractions (see below).

In what follows, we will concentrate on a special kind of persistent epistemic states
that will provide a representation for a non-prioritized belief change.

\begin{defn}
An epistemic state will be called \emph{persistent} if it satisfies
\begin{description}
\item[Persistence]\qquad If $s\prec t$, then $l(s)\subseteq l(t)$.
\end{description}
\end{defn}

For persistent epistemic states, the informational content of admissible belief states is
always preserved (persist) in transitions to more preferred states. This property
reflects the idea that the informational content of an admissible state is an essential
(though presumably not the only) factor in determining the place of this state in the
preference structure.

Persistent epistemic states are formally similar to a number of \emph{information models}
suggested in the logical literature, such as Kripke's semantics for intuitionistic logic,
Veltman's data semantics, etc. (see \cite{Ben88} for an overview). All such models
consist of a partially ordered set of informational states. The relevant partial order
represents possible ways of information growth, so it is assumed to satisfy the
persistence requirement by its very meaning.

As will be shown, persistent epistemic states constitute a smallest natural class of
epistemic states that is closed under belief change operations that do not involve
prioritization. Moreover, it is precisely this class of epistemic states that admits a
constructive representation in terms of flocks of bases.

An epistemic state will be called \emph{pure} if it satisfies

\begin{description}
\item[Pure Monotonicity]  $s\preceq t$ if and only if $l(s)\subseteq l(t)$.
\end{description}

Pure epistemic states is a special kind of persistent states for which the preference
relation is determined solely by the informational content of admissible belief states.
Accordingly, a pure epistemic state can be defined simply as a set of deductively closed
theories, with the intended understanding that the relation of set inclusion among such
theories plays the role of a preference relation. In other words, for pure epistemic
states, preference is given to maximal theories.

Pure epistemic states have been used in \cite{BASE,AGM} as a basis for a foundational
approach to belief change. As we mentioned, however, the approach has turned out to be
flawed, since it does not provide an adequate representation of belief expansions. The
present study stems from a more general approach to representing belief change suggested
in \cite{BOOK}.

\section{Bases and flocks}

Assume that a belief set $\B{B}$ is  generated by some base $\Delta$ with respect to a
certain consequence relation $\Th$ (that is, $\B{B}=\Th(\Delta)$). This structure is
representable as an epistemic state of the form $(P(\Delta),l,\prec)$, where admissible
belief states are the subsets of $\Delta$, $l$ assigns each such subset its deductive
closure, while $\prec$ is a preference relation on $P(\Delta)$. In the simplest
(non-prioritized) case, this preference relations is definable via set inclusion:
$\Gamma\prec\Phi$ iff $\Gamma\subset\Phi$.

In the latter case base-generated epistemic states are equivalent to pure epistemic
states consisting of theories of the form $\Th(\Gamma)$, where $\Gamma$ ranges over the
subsets of $\Delta$. Notice that any such epistemic state will be determinate, since it
contains a most preferred theory, namely $\Th(\Delta)$.

Unfortunately, we will see later that base-generated epistemic states (and bases
themselves) are arguably inadequate for representing belief contractions. A
generalization of bases that overcomes this shortcoming has been suggested in
\cite{FUKV86} and consists in using sets (or `flocks') of bases.

By a {\em flock\/} we will mean an arbitrary set of sets of propositions
$\C{F}=\{\Delta_i\}$, for $i\in I$. Such a flock can be considered as a collection of
bases $\Delta_i$, and the following construction of the epistemic state generated by a
flock can be seen as a natural generalization of base-generated epistemic states.

Any flock generates an epistemic state
$\BB{E}_{\C{F}}=\langle\C{F}_\downarrow,l,\prec\rangle$ defined as follows:
\begin{itemize}
\item $\C{F}_\downarrow$ is a set of all nonempty sets $\Gamma$ such that
$\Gamma\subseteq\Delta$, for some $\Delta\in\C{F}$;
\item $l(\Gamma)=\Th(\Gamma)$, for each $\Gamma\in\C{F}_\downarrow$;
\item $\Gamma\prec\Phi$ holds iff $\Gamma\subset\Phi$.
\end{itemize}

As can be seen, flocks constitute a generalization of bases. Namely, any base $\Delta$
can be identified with a singular flock $\{\Delta\}$.

As in our study, flocks were used in \cite{FUKV86} as a framework for belief change
operations. Our subsequent results will be different, however. The main difference can be
described as follows.

Let us say that two flocks are \emph{identical} if they generate the same epistemic
state. Now, let $\C{F}$ be a flock and $\Delta_0$ a set of propositions such that
$\Delta_0\subseteq\Delta$, for some $\Delta\in\C{F}$. Then it is easy to see that flocks
$\C{F}$ and $\C{F}\cup\{\Delta_0\}$ produce the same epistemic state, and consequently
they are identical in the above sense. This shows that a flock is determined, in effect,
by its inclusion-maximal elements. According to \cite{FUKV86}, however, the above two
flocks are distinct, and hence the validity of propositions with respect to a flock is
determined, in effect, by \emph{minimal} sets belonging to the flock. This makes the
resulting theory less plausible and more complex than it could be.

The above feature, though plausible by itself, gives rise, however, to high sensitivity
of flocks with respect to the syntactic form of the propositions occurring in them.

\begin{exam}
Let us consider the flock $\C{F}=\{\{A'\},\{A,B\}\}$, where $A'$ is logically equivalent
to $A$. Replacing $A'$ with $A$, we obtain a different flock $\C{F}'=\{\{A\},\{A,B\}\}$,
which is identical to $\{\{A,B\}\}$. Note that $A\land B$ is believed in the epistemic
state generated by the latter flock, though only $A$ is believed in the epistemic state
generated by the source flock $\C{F}$.
\end{exam}

The above example shows that flocks do not admit replacements of logically equivalent
propositions, at least in cases when such a replacement leads to identification of
propositions with other propositions appearing elsewhere in the flock. It should be kept
in mind, however, that the epistemic state generated by a flock is a syntax-independent
object, though purely syntactic differences in flocks may lead to significant differences
in epistemic states generated by them.

Flock-generated epistemic states are always persistent; this follows immediately from the
fact that $\Gamma\prec\Phi$ holds in such a state if and only if $\Gamma\subset\Phi$, and
hence $\Th(\Gamma)\subseteq\Th(\Phi)$. Moreover, it has been shown in \cite{BOOK} that
any finitary persistent epistemic state is representable by some flock. This means that
flocks constitute an adequate syntactic formalism for representing persistent epistemic
states. Unfortunately, this also shows that flocks are not representable by pure
epistemic states, as has been suggested in \cite{BASE}.

\section{Changing epistemic states}

Since belief sets are uniquely determined by epistemic states, operations on epistemic
states will also determine corresponding operations on belief sets. Two kinds of
operations immediately suggest themselves as most fundamental, namely removal and
addition of information to epistemic states.

\subsection{Contractions}

Admissible belief states of an epistemic state constitute all potential alternatives that
are considered as `serious possibilities' by the agent. In accordance with this, the
contraction of a proposition $A$ from an epistemic state $\BB{E}$ is defined as an
operation that removes all admissible belief states from $\BB{E}$ that support $A$. We
will denote the resulting epistemic state by $\BB{E}-A$.

The contraction operation has quite regular properties, most important of which being
\emph{commutativity}: a sequence of contractions can be performed in any order yielding
the same resulting epistemic state.

Let us compare the above  contraction with base contraction. According to \cite{Han93},
the first step in performing a base contraction of $A$ consists in finding preferred
(selected) subsets of the base that do not imply $A$. So far, this fits well with our
construction, since the latter subsets exactly correspond to preferred admissible
theories of the contracted epistemic state. Then our definition says, in effect, that the
contracted belief set should be equal to the intersection of these preferred theories.
Unfortunately, such a solution is unacceptable for the base paradigm, since we need to
obtain a unique contracted base; only the latter will determine the resulting belief set.
Accordingly, \cite{Han93} defines first the contracted base as the intersection of all
preferred subsets of the base (`partial meet base contraction'), and then the contracted
belief set is defined as the set of all propositions that are implied by the new base.

The problems arising in this approach are best illustrated by the following example
(adapted from \cite{Han92}).

\begin{exam}
Two equally good and reliable friends of a student say to her, respectively, that Niamey
is a Nigerian town, and that Niamey has a university. Our student should subsequently
retract her acquired belief that Niamey is a university town in Nigeria.
\end{exam}

Let $A$ and $B$ denote, respectively, propositions ``Niamey is a town in Nigeria", and
``There is a university in Niamey". Then the above situation can be described as a
contraction of $A\land B$ from the (belief set generated by the) base $\{A,B\}$. As has
been noted in \cite{GR95}, this small example constitutes a mayor stumbling block for the
base approach to belief change. Actually, we will see that none of the current approaches
handles satisfactorily this example.

To begin with, it seems reasonable to expect that the contracted belief set in the above
situation should contain $A\lor B$, since each of the acceptable alternatives support
this belief. This result is also naturally sanctioned by the AGM theory. Using the base
contraction, however, we should retreat first to the two sub-bases $\{A\}$ and $\{B\}$
that do not imply $A\land B$, and then form their intersection which happens to be empty!
In other words, we have lost all the information contained in the initial base, so all
subsequent changes should start from a scratch.

Next, it seems also reasonable to require that if subsequent evidence rules out $A$, for
example, we should believe that $B$. In other words, contracting first $A\land B$ and
then $A$ from the initial belief state should make us believe in $B$. This time the AGM
theory cannot produce this result. The reason is that the first contraction gives the
logical closure of $A\lor B$ as the contracted belief set, and hence the subsequent
contraction of $A$ will not have any effect on the corresponding belief state. Notice
that this information loss is not `seen' in one-step changes; it is revealed, however, in
subsequent changes.

As we see it, the source of the above problem is that traditional approaches to belief
change force us to choose in situations we have no grounds for choice. And our suggested
solution here amounts to retaining all the preferred alternatives as parts of the new
epistemic state, instead of transforming them into a single `combined' solution. This
means, in particular, that we should allow our epistemic states to be non-determinate.
This will not prevent us from determining each time a unique current set of beliefs; but
we should remember more than that.

Returning to the example, our contraction operation results in a new belief set
$\Th(A\lor B)$, as well as a new epistemic state $\{\Th(A),\Th(B)\}$ consisting of two
theories. This latter epistemic state, however, is not base-generated, though it will be
generated by a flock $\{\{A\},\{B\}\}$.

\subsubsection{Contraction of flocks}

Flocks emerge as a nearest counterpart of bases that will already be closed with respect
to our contraction operation. Actually, the latter will correspond to the operation of
deletion on flocks suggested in \cite{FUKV86}.

\begin{defn}
A \emph{contraction of a flock} $\C{F}=\{\Delta_i\}$ with respect to $A$ (notation
$\C{F}-A$) is a flock consisting of all maximal subsets of each $\Delta_i$ that do not
imply $A$.
\end{defn}

The following result confirms that the above operation on flocks corresponds to
contraction of associated epistemic states.

\begin{lem}
If $\BB{E}_{\C{F}}$ is an epistemic state generated by a flock $\C{F}$, then, for any
$A$, $\BB{E}_{\C{F}}-A=\BB{E}_{(\C{F}-A)}$.
\end{lem}

Despite the similarity of the above definition of contraction with that given in
\cite{FUKV86}, the resulting contraction operation will behave differently in our
framework. The following example illustrates this.

\begin{exam}
Contraction of the flock $\C{F}=\{\{A\},\{B\}\}$ with respect to $A$ is a flock
$\{\emptyset,\{B\}\}$, which is identical to $\{\{B\}\}$ according to our definition of
identity. Consequently, $B$ will be believed in the resulting epistemic state. This
behavior seems also to agree with our intuitions, since eliminating $A$ as an option will
leave us with $B$ as a single solution. In the representation of \cite{FUKV86}, however,
$\{\emptyset,\{B\}\}$ is reducible to $\{\emptyset\}$. Consequently, nothing will be
believed in the resulting epistemic state. Furthermore, this also makes the corresponding
operation of deletion \emph{non-commutative}: whereas deletion of $A\land B$ and then $A$
from the base $\{A,B\}$ results in $\{\emptyset,\{B\}\}$, deletion of $A$ first and then
$A\land B$ will give a different flock, namely $\{\{B\}\}$.
\end{exam}

\subsection{Merge and expansion}

The operation of expansion consists in adding information to epistemic states. In the AGM
theory, this is achieved through a straightforward addition of the new proposition to the
belief set, while in the base approach the new proposition is added directly to the base.

The framework of epistemic states drastically changes, however, the form and content of
expansion operations. This stems already from the fact that adding a proposition to an
epistemic state is no longer reducible to adding it to the belief set; it should
determine also the place of the newly added proposition in the structure of the expanded
epistemic state. This will establish the degree of firmness with which we should believe
the new proposition, as well as its dependence and justification relations with other
beliefs. As a way of modelling this additional information, we suggest to treat the
latter as a special case of merging the source epistemic state with another epistemic
state that will represent the added information. Accordingly, we will describe first some
merge operation on epistemic states. Then an expansion will be defined roughly as a merge
of $\BB{E}$ with a rudimentary epistemic state that is generated by the base $\{A\}$.

\subsubsection{Merging epistemic states}

Merge is a procedure of combining a number of epistemic states into a single epistemic
state, in which we seek to combine information that is supported by the source epistemic
states. It turns out that this notion of merging can be captured using a well-known
algebraic construction of \emph{product}. Roughly, a merge of two epistemic states
$\BB{E}_1$ and $\BB{E}_2$ is an epistemic state in which the admissible states are all
\emph{pairs} of admissible states from $\BB{E}_1$ and $\BB{E}_2$, the labelling function
assigns each such pair the deductive closure of the union of their corresponding labels,
while the resulting preference relation agrees with the `component' preferences (see
\cite{BOOK} for a formal description).

Since the primary subject of this study is non-prioritized change, we will consider below
only one kind of merge operations, namely a pure merge that treats the source epistemic
states as having an equal `weight'.

\begin{defn}
A \emph{pure merge} of epistemic states $\BB{E}_1=(\C{S}_1,l_1,\prec_1)$ and
$\BB{E}_2=(\C{S}_2,l_2,\prec_2)$ is an epistemic state
$\BB{E}_1\times\BB{E}_2=(\C{S}_1\times\C{S}_2,l,\prec)$ such that

\begin{itemize}
\item $l(s_1,s_2)=\Th(l_1(s_1)\cup l_2(s_2))$, for any
$(s_1,s_2)\in\C{S}_1\times\C{S}_2$;

\item $(s_1,s_2)\preceq(t_1,t_2)$ iff $s_1\preceq_1 t_1$ and $s_2\preceq_2 t_2$.
\end{itemize}
\end{defn}

A pure merge is a merge operation that treats the two component epistemic states as two
equally reliable sources. It is easy to see, in particular, that it is a commutative
operation.

Being applied to base-generated epistemic states, pure merge corresponds to a
straightforward union of two bases:

\begin{lem}
If $\BB{E}_1$ and $\BB{E}_2$ are epistemic states generated, respectively, by bases
$\Delta_1$ and $\Delta_2$, then $\BB{E}_1\times\BB{E}_2$ is equivalent to an epistemic
state generated by $\Delta_1\cup\Delta_2$.
\end{lem}

\subsubsection{Merging flocks}

To begin with, the following result shows that pure merge preserves persistence of
epistemic states.

\begin{lem}
A pure merge of two persistent epistemic states is also a persistent epistemic state.
\end{lem}

Since finitary persistent epistemic states are representable by flocks, a pure merge
gives rise to a certain operation on flocks. This operation can be described as follows:

Let us consider two flocks $\C{F}_1$ and $\C{F}_2$ that have no propositions in common.
Then a \emph{merge} of $\C{F}_1$ and $\C{F}_2$ will be a flock
\[\C{F}_1\times \C{F}_2=\{\Delta_i\cup\Delta_j\mid \Delta_i\in\C{F}_1,\Delta_j\in
\C{F}_2\}\]

Thus, the merge of two disjoint flocks is obtained by a pairwise combination of bases
belonging to each flock. Note, however, that the assumption of disjointness turns out to
be essential for establishing the correspondence between merge of flocks and pure merge
of associated epistemic states. Still, this requirement is a purely syntactic constraint
that can be easily met by replacing some of the propositions with logically equivalent,
though syntactically different propositions. A suitable example will be given later when
we will consider expansions of flocks that are based on the above notion of merge.

The following result shows that merge of flocks corresponds exactly to a pure merge of
associated epistemic states.

\begin{thm}
If $\BB{E}_{\C{F}}$ and $\BB{E}_{\C{G}}$ are epistemic states generated, respectively, by
disjoint flocks $\C{F}$ and $\C{G}$, then $\BB{E}_{\C{F}}\times\BB{E}_{\C{G}}$ is
isomorphic to $\BB{E}_{\C{F}\times \C{G}}$.
\end{thm}

\subsubsection{Pure expansions}

For any proposition $A$, we will denote by $\BB{E}_A$ the epistemic state generated by a
singular base $\{A\}$. This pure epistemic state consists of just two theories, namely
$\Th(\emptyset)$ and $\Th(A)$. Accordingly, it gives a most `pure' expression of the
belief in $A$. Now the idea behind the definition below is that an expansion of an
epistemic state with respect to $A$ amounts to merging it with $\BB{E}_A$.

\begin{defn}
A \emph{pure expansion} of an epistemic state $\BB{E}$ with respect to a proposition $A$
(notation $\BB{E}+A$) is a pure merge of $\BB{E}$ and the epistemic state $\BB{E}_A$ that
is generated by a base $\{A\}$.
\end{defn}

In a pure expansion the new proposition is added as an independent piece of information,
that is, as a proposition that is not related to others with respect to priority. Being
applied to base-generated epistemic states, pure expansion corresponds to a
straightforward addition of a new proposition to the base.

\begin{cor}
If $\BB{E}$ is generated by  a base $\Delta$, then $\BB{E}+A$ is isomorphic to an
epistemic state generated by $\Delta\cup\{A\}$.
\end{cor}

Since pure merge is a commutative operation, pure expansions will also be commutative.

As any other kind of change in epistemic states, expansions generate corresponding
changes in belief sets of epistemic states. It turns out that expansions generate in this
sense precisely AGM belief expansion functions:

\begin{lem}
If $\B{B}$ is a belief set of $\BB{E}$, then the belief set of $\BB{E}+A$ coincides with
$\Th(\B{B},A)$.
\end{lem}

Thus, belief expansions generated by expansions of epistemic states behave just as AGM
expansions: the underlying epistemic state plays no role in determining the resulting
expanded belief set, since the latter can be obtained by a direct addition of new
propositions to the source belief set. It should be kept in mind, however, that identical
expansions of belief sets can be produced by expansions of different epistemic states,
and even by different expansions of the same epistemic state. These differences will be
revealed in subsequent contractions and revisions of the expanded belief set.

\subsubsection{Expansions of flocks}

As we have seen earlier, pure merge generates a corresponding merge operation on flocks.
Consequently, pure expansion corresponds in this sense to a certain expansion operation
on flocks.

\begin{defn}
An \emph{expansion of a flock} $\C{F}=\{\Delta_i\mid i\in I\}$ with respect to a
proposition $A$ that does not appear in $\C{F}$ is a flock
$\C{F}+A=\{\Delta_i\cup\{A\}\mid i\in I\}$.
\end{defn}

Thus, an expansion of a flock is obtained simply by adding the new proposition to each
base from the flock. Our earlier results immediately imply that this operation exactly
corresponds to a pure expansion of the associated epistemic state with $A$:

\begin{cor}
If $\BB{E}_{\C{F}}$ is an epistemic state generated by the flock $\C{F}$, and $A$ does
not appear in $\C{F}$, then $\BB{E}_{\C{F}}+A$ is isomorphic to $\BB{E}_{\C{F}+A}$.
\end{cor}

The above operation is quite similar in spirit to the operation of insertion into flocks
used in \cite{FUKV86}, though the latter was intended to preserve consistency of the
component bases, so they defined, in effect, the corresponding revision operation based
on contraction and expansion in our sense. Note, however, that our flock expansion is
defined only when the added proposition does not appear in the flock. The need for the
restriction is illustrated by the following example.

\begin{exam}
Let us return to the flock $\C{F}=\{\{A\},\{B\}\}$, where $A$ and $B$ denote,
respectively, propositions ``Niamey is a town in Nigeria", and ``There is a university in
Niamey". Recall that this flock is obtainable by contracting $A\land B$ from the base
$\{A,B\}$. In other words, it reflects an informational situation in which we have
reasons to believe in each of these propositions, but cannot believe in both.

Now let us expand the epistemic state $\BB{E}_{\C{F}}$ with $B$. This expansion can be
modeled by expanding $\C{F}$ with some proposition $B'$ that is logically equivalent to
$B$. In other words, the epistemic state generated by the flock
$\C{F}+B'=\{\{B',A\},\{B',B\}\}$ will be equivalent to the expansion of $\BB{E}_{\C{F}}$
with $B$. This flock sanctions belief in $B$ in full accordance with our intuitions.
Actually, it can be shown that the latter flock is reducible to a flock
$\{\{B',A\},\{B\}\}$ in the sense that the latter flock will produce an equivalent
epistemic state. However, $B'$ cannot be replaced with $B$ in these flocks: the flock
$\{\{B\},\{B,A\}\}$ is already reducible to a single base $\{\{A,B\}\}$ in which both $A$
and $B$ are believed, contrary to our intuitions about the relevant situation: receiving
a new support for believing that there is a university in Niamey should not force us to
believe also that Niamey is a town in Nigeria. This also shows most vividly that a
straightforward addition of $B$ to each base in the flock $\{\{A\},\{B\}\}$ does not
produce intuitively satisfactory results\footnote{Exactly this was a by-product of the
representation suggested in \cite{BASE}.}.

An additional interesting aspect of the above representation is that, though we fully
believe in $B$ in the flock $\{\{B\},\{B',A\}\}$, the option $A$ has not been forgotten;
if we will contract now $B$ from the latter flock, we will obtain the flock $\{\{A\}\}$
which supports belief in $A$. A little reflection shows that this is exactly what would
be reasonable to believe in this situation.
\end{exam}

\section{Conclusions}

The purpose of this study was to give a formal representation for iterative
non-prioritized change. As has been shown, such a representation can be achieved in the
framework of persistent epistemic states, with flocks providing the corresponding
constructive representation.  Moreover, these representations overcome shortcomings of
both the AGM and base-oriented models that  incur loss of information in iterative
changes.

Contraction and expansions are two basic operations on epistemic states that allow to
define the majority of derived belief changes. Thus, a \emph{revision} of an epistemic
state is definable via Levi identity (on the level of epistemic states), namely as a
combination of contraction and expansion. As can be shown, the resulting operation will
be sufficiently expressive to capture any relational belief revision function in the
sense of AGM.

To conclude the paper, we want to mention an interesting problem concerning expressivity
of our belief change operations vis-a-vis flocks. Namely, it has been shown in
\cite{BOOK} that there are flocks that are not constructible from simple primitive flocks
using the contraction and expansion operations (a simplest example being the flock
$\{\{p\},\{p\land q\}\}$). This apparently suggests that our stock of belief change
operations is not complete and need to be extended with other operations that would
provide functional completeness with respect to constructibility of flocks. The resulting
theory would give then a truly complete constructive representation of non-prioritized
belief change.

\bibliographystyle{named}

\end{document}